\documentclass[letterpaper]{article} 
\usepackage{aaai25}  
\usepackage{times}  
\usepackage{helvet}  
\usepackage{courier}  
\usepackage[hyphens]{url}  
\usepackage{graphicx} 
\usepackage{natbib}  
\usepackage{caption} 
\usepackage{algorithm}
\usepackage{algorithmic}

\usepackage{newfloat}
\usepackage{listings}
\DeclareCaptionStyle{ruled}{labelfont=normalfont,labelsep=colon,strut=off} 
\floatstyle{ruled}
\newfloat{listing}{tb}{lst}{}
\floatname{listing}{Listing}
\usepackage{markdown}
\usepackage{soul}
\usepackage{tabularx}
\usepackage{textcomp}

\title{Enhanced Question-Answering for Skill-based learning using Knowledge-based AI and Generative AI}

\author {
    Rahul K. Dass,
    Rochan H. Madhusudhana,
    Erin C. Deye,
    Shashank Verma,
    Timothy A. Bydlon,
    Grace Brazil,
    Ashok K. Goel
}
\affiliations {
    Georgia Institute of Technology \\
    \{rdass7,rochan.hm,edeye3,sverma342,tbydlon3,gbrazil2\}@gatech.edu, ashok.goel@cc.gatech.edu
}

\begin{document}

\maketitle

\begin{abstract}
Supporting learners’ understanding of taught skills in online settings is a longstanding challenge.
While exercises and chat-based agents can evaluate understanding in limited contexts, this challenge is magnified when learners seek explanations that delve into procedural knowledge (\textit{how} things are done) and reasoning (\textit{why} things happen).
We hypothesize that an intelligent agent's ability to understand and explain learners' questions about skills can be significantly enhanced using the TMK (Task-Method-Knowledge) model, a Knowledge-based AI framework.
We introduce \textit{Ivy}, an intelligent agent that leverages an LLM and iterative refinement techniques to generate explanations that embody teleological, causal, and compositional principles.
Our initial evaluation demonstrates that this approach goes beyond the typical shallow responses produced by an agent with access to unstructured text, thereby substantially improving the depth and relevance of feedback. 
This can potentially ensure learners develop a comprehensive understanding of skills crucial for effective problem-solving in online environments.
\end{abstract}

%

\section{Introduction}

Online education platforms such as MOOCs and professional certification programs have significantly expanded access to education.
Despite extensive literature on effective online course design \cite{kay2012exploring,hansch2015video}, these platforms often inadequately support \textit{skill-based learning} \cite{squire_skill_learning, jeremy_skill_learning}, particularly when learners require in-depth explanations on procedural knowledge and reasoning.
Addressing this gap is crucial as the demand for reskilling and upskilling grows in today's rapidly evolving job market \cite{WEF2019,goel2024ai}.

Traditional online tools like instructional videos and exercises typically promote passive learning \cite{chi2014icap,chi2018icap}.
Our research extends recent work \cite{goel2016design,goel2017using,ou2016designing,ou2019designing} that have demonstrated how to effectively incorporate pedagogical and instructional strategies to foster active forms of learning, thereby enhancing the understanding of taught skills and cognitive engagement \cite{chi2014icap,chi2018icap}.

Recent technological advances, including Intelligent Agents (IAs) \footnote{While the development of intelligent agents predates the rise of Generative AI, in this paper, we refer to Ivy—an LLM-powered intelligent agent designed to support skill-based learning.} developed using Large Language Models (LLMs), have been integrated into online platforms \cite{jurenka2024towards,taneja2024jill,chevalier2024language,kakar2024jill}. 
Yet, LLM-based agents tend to deliver general or ``shallow'' understanding and struggle with the deep, procedural understanding required for effective skill-based learning \cite{kambhampati2021language,valmeekam2022large,ruis2024procedural}.
We argue that such agents, relying on unstructured text, fail to address learners’ deeper questions about ``how'' and ``why'' skill-based tasks or mechanisms are performed.

To address these limitations, we propose a hybrid approach that integrates Knowledge-based AI (KBAI)--representing skills using a structured framework called TMK (Task-Method-Knowledge) \cite{murdock2008meta,rugaber2013gaia,goel2017gaia}--with Generative AI to enhance Ivy, an intelligent agent designed to generate explanations to learners' questions about skills taught in a graduate-level online AI course.
This builds on previous research that combined similar integrations for generating self-explanations by AI agents across various educational settings \cite{basappa2024social,sushri2024combining}.
While Ivy is envisioned to be embedded within videos, its current implementation operates as a standalone question-answering system.

Our research explores the following key research questions (RQs) and research hypotheses (RHs):
\begin{description}
    \item[RQ1:] How can Ivy explain how a skill functions?
    \begin{description}
        \item[RH1:] Ivy can explain a skill's functionality by utilizing a structured framework, like a TMK model, to represent the skill’s design and leverages Generative AI to provide introspective responses to learners’ queries.
    \end{description}
    \item[RQ2:] How can Ivy inspect the design of a skill?
    \begin{description}
            \item[RH2.1] A TMK model organizes a skill's components and procedures into Task, Method and Knowledge modules.
            \item[RH2.2] A TMK model embodies teleological principles through explicit Task-Method linkages and captures causality via state sequences and transitions within Methods.
    \end{description}
\end{description}

In this paper, we make four contributions. 
(1) We detail how we represent skills as TMK models from an online AI course.
(2) We outline Ivy's architecture\footnote{Developed using GPT-4o-mini and LangChain \cite{langchain2023}} that dynamically generates explanations for skill-based learning questions.
(3) We implement a comprehensive evaluation strategy using human-centric Question-Answering (QA) and automated metrics to validate the Ivy’s explanatory capabilities.
(4) By comparing responses from TMK models with those generated from the AI course’s textbook\footnote{Video transcripts of the online AI course.}, we show our approach significantly improves the agent’s ability to deliver meaningful explanations, enhancing skill-based learning in online education.

\section{Related Work}

\subsection{Skill-Based Learning and its Representation in Intelligent Agents}

Skill-based learning focuses on developing cognitive abilities crucial for problem-solving within educational contexts.
It involves transitioning from declarative knowledge (understanding facts and principles) to procedural knowledge (applying skills).
The ACT-R Theory \cite{anderson1983architecture} describes this transition, emphasizing the shift from concept-based learning (``what knowledge'') to skill-based learning (``how to'' and ``why'' knowledge) \cite{bransford2000people,ryle2009concept}.
Effective skill acquisition requires an understanding of the theoretical principles and applying them in varied situations.

The representation of skills in intelligent agents plays a critical role in adaptive learning systems.
Traditional systems like Cognitive Tutors \cite{anderson1995cognitive,koedinger2006cognitive,rau2009intelligent} model skills using rule-based formulations based on the ACT* theory of learning and problem solving \cite{anderson1983architecture,anderson1993rules}.
These tutors have facilitated skill-based learning  in domains like programming skills in LISP \cite{anderson1984learning}, geometry \cite{anderson1981acquisition}, and fractions \cite{rau2009intelligent}.
Despite their success, such systems often struggle with providing deep explanations queries and scalability.
Alternative approaches, such as ontology-based frameworks, offer structured representations of cognitive skills, particularly in K-12 settings \cite{cogskillnet}.

\subsection{Question-Answering using AI}
Recent advancements in AI-driven question-answering (QA) using AI have leveraged deep learning and transformer-based architectures like BERT \cite{devlin2018bert} and GPT, enhancing retrieval-augmented systems for knowledge-based QA (KBQA). 
These developments integrate structured knowledge bases and natural language processing techniques to improve query handling and answer generation \cite{knowledge_repr_qa,moldovan2002lcc,chu2003question,tari2005using}.

For instance, Braz et al. (2005) employed a hierarchical knowledge representation called EFDL (Extended Feature Description Logic) and used Integer Linear Programming and phrase-level subsumption algorithms to generate answers to existing QA databases \cite{de2005knowledge}.
Balduccini et al. (2008) converted English text to logical representation and then used automated logical theorem provers to extract facts and answer questions \cite{balduccini2008knowledge}.

The incorporation of LLMs has further advanced KBQA, introducing few-shot learning capabilities and complex question handling and refined answer retrieval \cite{chen2021retrack,tan2023can}.
Despite these innovations, there is still a need to harness these technologies more effectively for modeling dynamic problem-solving skills within online learning environments, potentially transforming explanation generation in educational settings.

\subsection{Generating Reasoned Responses using LLMs}
While LLMs excel at retrieval tasks and text completion using methods like retrieval augmented generation (RAG) \cite{lewis2020retrieval}, their capability for structured reasoning, such as planning or tasks requiring logical consistency, remains limited.
Critics suggest that LLMs, often acting as enhanced n-gram models, rely too much on pattern recognition, leading to responses that appear reasoned but lack depth \cite{valmeekam2023planning,kambhampati2024can}.

To enhance LLM reasoning, methods like Chain-of-Thought (CoT) prompting have been introduced to bolster LLM reasoning \cite{cot}.
However, the effectiveness of CoT prompting can vary greatly depending on the task complexity and model scale, occasionally leading to model overfitting \cite{stechly2024chain}.

Building on these insights, our approach employs a TMK decomposition to create a structured knowledge base for problem-solving.
By integrating this framework with Ivy, a GPT 4-based agent enhanced by LangChain and prompt engineering, we enable it to generate responses that demonstrate both procedural accuracy and a deep understanding of teleological and causal dimensions.

\section{Methodology}

\subsection{Modeling Skills using TMK}
\begin{figure}[t]
  \centering
  \includegraphics[scale=0.75]{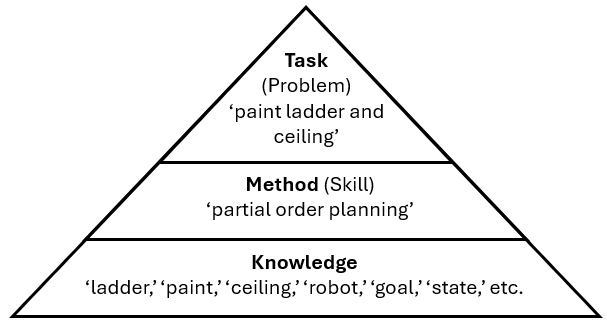}
  \caption{High-level TMK model of the `Partial Order Planning' skill, showing hierarchical problem decomposition.}
  \label{fig:ivy-tmk}
\end{figure}

As stated in RQ1 and RH1, we posit that, for Ivy to effectively explain how a skill functions, skills must be modeled using the TMK (Task-Method-Knowledge) framework \cite{murdock2008meta,rugaber2013gaia}.
While other frameworks such as BDI (Belief-Desire-Intention) \cite{rao1997modeling} and HTN (Hierarchical Task Network) \cite{ghallab2004automated} focus on modeling an agent's goals and actions, they have not been specifically applied to represent skills in online courses or for an AI agent to introspectively generate detailed explanations.

In this paper, we modeled six skills\footnote{A total of 22 lessons or skills are taught in the AI course.} from an online AI course offered in a master's program in computer science at Georgia Institute of Technology which has been offered since Fall 2014.
The skills were manually modeled using the TMK framework by four graduate research assistants and underwent a meticulous peer review process to ensure accuracy and consistency.

\subsubsection{TMK Model Development Steps}
To create a TMK model, a sequence of steps were followed:
\begin{itemize}
  \item \textbf{Task Definition}: Identify the goal of a skill, allowable \texttt{inputs}, \texttt{givens} (pre-conditions), \texttt{makes} (post-conditions), and \texttt{outputs}.
  \item \textbf{Method Specification}: Outline the sequence of states and state transitions to accomplish the task, using an \texttt{Organizer} (deterministic finite state machine). 
  For example, in a sorting algorithm, a method might involve fixing the order of two indices in a list by performing steps such as comparing the pairs of numbers and swapping them if they are out of order.  
  \item \textbf{Knowledge Representation}: Define objects, concepts, and their relationships within the environment, including the properties of the objects and the logical expressions that connect with user-supplied values.
  Going back to the sorting algorithm example, the knowledge representation may include the concepts of numbers, lists, and the relationships between them.
  
  \item \textbf{Hierarchical Aspect of TMKs}: \textbf{Tasks} can be decomposed into hierarchical sub-goals through their \textbf{Method} specification. In the Method specification, each state can have a sub-goal, allowing a TMK structure to model an arbitrarily complex skill.
\end{itemize}

Fig. \ref{fig:ivy-tmk} shows a high-level TMK model for the `Partial Order Planning skill', used in a `robot painting a ladder and ceiling' problem, demonstrating the application of skills taught in the AI course.
The Task component aligns with the problem (``paint ladder and ceiling'') and is connected to the Method component (``partial order planning'').
As stated in RH2, this part of the model illustrates two characteristics critical for generating procedural explanations: (1) the linkage between Tasks and Methods showcases teleological principles, where goals (tasks) systematically determine the methods used; and (2) the sequence of states and state transitions within Methods capture causality.
The Knowledge component includes concepts with properties and ground truths, which are essential for executing the Method and completing the Task.

\subsection{Generating Reasoned Responses using TMK Models}
In skill-based learning, Ivy utilizes the TMK model to facilitate teleological, causal, and compositional reasoning.
This allows it to generate responses that comprehensively map procedural steps to their underlying skill objectives and ultimately, its overarching goal.
For instance, in the context of the ``Classification'' skill, Ivy follows a sequence of state transitions to classify an object, such as a bird.
These steps include: (1) Processing percepts: Identifying observable features (e.g., wings, beak), (2) Mapping percepts to equivalence classes: Grouping features into predefined categories (e.g., “avian features”), (3) Classifying objects: Assigning the object to a specific class (e.g., “bird”), (4) Validating the classification: Confirming the assignment aligns with the criteria.

This sequence enables Ivy to explicitly represent causal relationships between each step.
Furthermore, by aligning these steps to the ``Classification'' method, Ivy showcases a teleological connection, demonstrating how the procedural sequence serves the broader goal of classification.
Additionally, the hierarchical structure of TMK models equips Ivy to break down complex tasks into simpler, actionable subtasks.
This compositional aspect ensures that Ivy not only explains ``what'' needs to be done, but also provides insight into ``why'' each step is necessary and ``how'' it contributes to the overall skill.

Therefore, by integrating the TMK model with Generative AI, Ivy can go beyond procedural steps to explain the underlying reasoning and organizational logic of a skill, fostering deeper learner comprehension.

\subsection{End-to-End Architecture of Ivy}
The end-to-end architecture of Ivy—from classifying a learner's question to accessing relevant TMK components, generating a knowledge trace and sending the learner an output—is outlined in Fig. \ref{fig:ivy-schematic-flow}.
Initially, Ivy assesses whether a learner’s question is \textit{relevant} by checking for semantic similarities between the question’s keywords and the top-level names of Tasks, Methods, and Knowledge components within the skill's TMK model for the lesson that the learner is currently undertaking.
This is achieved through zero-shot classification using an LLM.

\subsubsection{Knowledge Retrieval Module}

\begin{figure}[htb]
  \centering
  \includegraphics[scale=0.5]{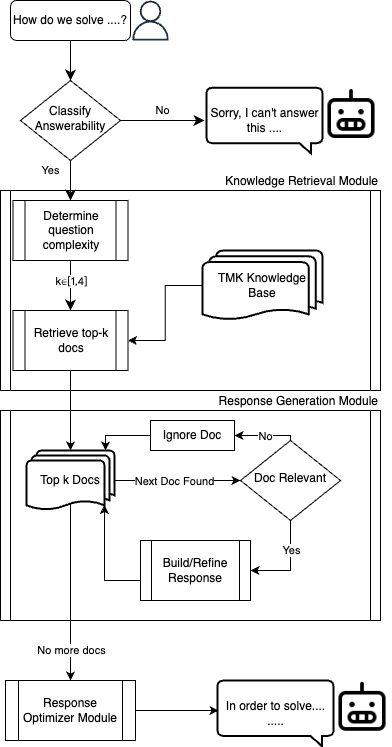}
  \caption{Overall schematic of Ivy's architecture.}
  \label{fig:ivy-schematic-flow}
\end{figure}

This stage of Ivy focuses on retrieving information to generate responses that clarify the relationships and operational mechanisms within the skill framework. 
As shown in Fig. \ref{fig:ivy-schematic-flow}, Ivy identifies the most relevant Task, Method, and Knowledge components to address the learner’s query, ensuring responses are both accurate and contextually meaningful. This process involves two key steps:

\begin{enumerate}
    \item \emph{Question Complexity Assessment:} We prompt Ivy to assess the complexity and level of detail expected by the learner's question, see Appendix \ref{appendix:question-complex-prompt} for a detailed description of the prompt.
    A \(\texttt{k-score} \in [1, 4]\) is assigned to determine the number of documents needed from the skill's TMK model for a suitable response.
    A k-score of 1 calls for a brief response (3-5 words), while a score of 4 indicates a comprehensive response spanning multiple paragraphs sourced from top-4 matched documents.
    This scoring helps Ivy tailor the depth and breadth of information included in its responses.
    \item \emph{Document Retrieval:} Based on the \texttt{k-score}, Ivy uses the FAISS library \cite{faiss} to retrieve the top $k$ relevant documents from the skill’s TMK model.
    FAISS efficiently searches and clusters dense vector embeddings, capturing deeper semantic relationships beyond traditional text matching.
    For example, it recognizes the similarity between terms like `ship' and `boat', or the parallel between `man' and `woman' to `king' and `queen'.
    This capability enables Ivy to identify the most relevant content, ensuring responses are both precise and educationally impactful.
\end{enumerate}

\subsubsection{Response Generation Module}
Once the relevant documents are retrieved, Ivy constructs its response through an iterative refinement process:
\begin{enumerate}
    \item \emph{Initial Response Generation}: Ivy creates an initial response using the top-most relevant document from the retrieved set.
    \item \emph{Response Refinement}:  It then sequentially refines this response by incorporating information from the remaining $k-1$ documents.
\end{enumerate}

This iterative process improves the response’s accuracy, completeness, and relevance by adding context and details that may not have been covered in the initial document.


\subsubsection{Response Optimizer Module}
After generating a response from the top $k$ relevant documents, Ivy refines and finalizes it before delivering it to the user. During testing, we observed that LLMs often produce verbose or repetitive content when explaining concepts, occasionally including phrases like ``based on the previous information...'' that reveal the iterative refinement process.
To address this, Ivy uses a \textit{response optimizer} to rephrase and adjust the response’s verbosity based on the question’s intent, ensuring clarity and conciseness. For questions requiring detailed explanations, such as ``how’’-based procedural queries, the optimizer retains depth, using paragraphs or bullet points for clarity. In contrast, simpler queries, like definitions, are distilled into concise responses no longer than two lines.
While this module’s logic aligns with the ``Question Complexity Assessment'' stage, which determines the number of documents to retrieve based on the \texttt{k-score}, its role here is to ensure that the final response is clear, actionable, and tailored to the user’s needs. This step acts as a final safeguard, balancing thoroughness with readability.

\section{Experimental Design and Evaluation Strategies}
To evaluate Ivy's ability to generate responses to skill-based learning questions, we conducted a comprehensive internal evaluation inspired by recent advancements in human-centered QA and automated metrics \cite{jurenka2024towards}.
First, we developed a diverse set of verification questions to evaluate every component of the TMK model.
Next, we compared Ivy's responses to those generated by baseline models using a preference voting system.
This evaluation was guided by ``explanation quality properties'' from the Explanation AI (XAI) literature \cite{nauta2023anecdotal}.
We refer to this process as the ``Developer Perception Evaluation''.
Secondly, for automated evaluation, we employed independent LLM-based judges to inspect Ivy's knowledge trace and verify the accuracy and relevance of responses based on three distinct criteria (described in section \ref{sec:automated-eval}).
Lastly, we used semantic similarity scores using embeddings \cite{reimers-2019-sentence-bert} to compare Ivy's generated response with the ``expected answers''.

\subsection{Developer Perception Evaluation with Baseline model}
\subsubsection{Verification Questions}
We designed five categories of evaluation questions specific to aspects of skill-based learning and evaluated Ivy using a total of 30 questions across these categories, covering six skills. 
See Appendix A.1 for the complete list.
\begin{enumerate}
    \item \emph{Task Questions}: focus on the goal or achievable objectives using the skill. 
    Example: ``What is the goal of solving a block world problem with means-end analysis?'' (Skill: Means-End Analysis).
    \item \emph{Method Questions:} inquire about procedural steps needed to complete a task. 
    Example: ``What is required to map percepts to equivalence classes?'' (Skill: Classification).
    \item \emph{Knowledge Questions:} focus on foundational information like terms or definitions related to the skill. 
    Example: ``What is a literal in the context of logical sentences?'' (Skill: Logic).
    \item \emph{Student Questions:} are learner specific queries agnostic of modeling design choices and expected to mimic a real student's questions while learning the skill.
    Example: ``What make a state 'productive' in addition to being `legal'?'' (Skill: Semantic Networks).
    \item \emph{Cannot Answer Questions:} are out-of-scope to the skill's domain to assess how Ivy balances precision with recall.
    Example: ``How do you make a quesadilla?''.
\end{enumerate}

\subsubsection{RAG Benchmark Model}
To validate our hypotheses RH1 and RH2, we compared Ivy with a RAG-based model built on GPT-4\footnote{Similar to Ivy, we used GPT-4o mini}, referred to as the ``RAG Benchmark’’ which uses the AI course textbook.

Both systems were provided with identical base system prompts to ensure a fair comparison based on their respective KBAI and Generative AI capabilities. The RAG Benchmark, although not integrated with LangChain, used a standard OpenAI Assistant augmented with a file search tool that queried a vector store of the AI course textbook embeddings— an approach commonly seen in modern educational AI applications.
This comparative analysis aimed to evaluate whether the TMK-based backend provides a significant advantage in generating more meaningful and contextually relevant responses to skill-based learning questions, compared to a model relying on RAG with unstructured course materials.

\subsection{Semantic Similarity of Ivy's Responses with Developer-Expected Responses}
We utilized SBERT (Sentence-BERT) \cite{reimers-2019-sentence-bert}, to compute sentence embeddings for effective semantic similarity assessment.
For each verification question, we created an \textit{expected response} by manually tracing through the TMK models they had developed.
To evaluate Ivy’s performance, we compared the embeddings of the \textit{Ivy’s response} with those of the \textit{expected response} using SBERT. The resulting scores ranged from 0 to 1, where 0 indicates no similarity and 1 indicates a high degree of semantic similarity between the two responses.

\subsection{Automated Evaluation of Knowledge Traces using LLM Judges}
\label{sec:automated-eval}
To ensure Ivy adheres to its design principles, we implemented an automated knowledge trace evaluation system using LLM as ``judges'' to assess how Ivy uses TMK files during response generation, inspired by recent work \cite{zheng2023judging,ke2024critiquellm}.
The evaluation focused on three criteria: (1) whether Ivy accessed the most relevant TMK documents for each question, (2) to what extent the intermediate response (from Response Generation Module) relied on these documents versus hallucinated content, and (3) whether critical information was maintained in the final responses after optimization by the Response Optimizer Module.

Ivy was configured to log its knowledge trace as it processed questions, storing all data necessary for analyzing the three criteria. 
To counteract the inherent stochastic nature of LLMs, we processed each of the 30 verification questions five times, resulting in 150 responses and corresponding knowledge traces.

Two automated LLM-based judges analyzed these responses.
By providing the TMK model, learner's question, and the generated final response as context, the first judge evaluated how much of the intermediate response content was directly derived from the accessed TMK files, using Chain-of-Thought reasoning \cite{cot} to document its analysis process and identify any externally generated content.
The second judge assessed the retention of information from the intermediate response to the final response, noting any omissions or alterations and providing detailed rationales for their findings. 
To ensure consistency and quality, we audited a random sample of these evaluations.

\section{Results}
\subsection{Developer Perception Evaluation}
Seven evaluators from the Ivy development team used a set of 30 verification questions to compare responses generated by Ivy and the RAG Benchmark, focusing on metrics such as correctness, completeness, confidence, comprehensibility, and compactness \cite{jurenka2024towards}\footnote{See Appendix \ref{appendix:xai-metrics} for descriptions regarding each metric and desired outcomes for human-centered evaluations of generated AI-based explanations.}.
Despite using internal developers as evaluators, this setup gave preliminary insights into how the responses might be perceived by real-world users--adult learners engaged in online courses.

While we captured ratings across all five metrics, developers were also asked to provide an overall preference as a vote when reviewing a question alongside the two generated responses.
Although a detailed analysis of developers’ ratings per metric is beyond the scope of this paper, we use the developers’ overall votes as a proxy for these metrics and focus on reporting these findings in our results.

To minimize bias, evaluators conducted blind voting, where they did not know which system generated each response. They could choose one response, both responses, or neither as the preferred option.
The voting outcomes indicated a preference for Ivy, which received 115 votes compared to the RAG Benchmark’s 75 out of 140 total evaluations.
This suggests that Ivy’s responses, guided by the TMK framework, were deemed more helpful by evaluators. 
The average agreement indices (percentage of total responses from a system marked as ideal by evaluators) for the RAG Benchmark and Ivy were 53.57\% and \textbf{82.14\%} respectively, reflecting a stronger consensus among developers for the Ivy’s responses.
A detailed voting results and breakdowns per question-category and skill are presented in Appendices \ref{appendix:results-category} and \ref{appendix:results-skill}.
\begin{table}[h!]
    \centering
    \begin{tabular}{|l|c|c|} \hline 
         \rule{0pt}{2ex}\textbf{Backend} & \textbf{Total Votes}  & \textbf{Agreement Index}\\ \hline 
         \rule{0pt}{2ex}RAG Benchmark&  75& 53.57\%\\ \hline 
         \rule{0pt}{2ex}\textbf{Ivy}&  115& \textbf{82.14\%}\\ \hline 
    \end{tabular}
    \caption{Developer perception evaluation using 30 diverse verification questions between RAG Benchmark and Ivy.}
    \label{tab:overall_results}
\end{table}

A limitation of human-centered evaluations for LLM-generated responses is the potential variation in responses evaluators may encounter due to the stochastic nature of LLMs, even when the temperature is set to 0. However, an analysis of stored responses showed minimal differences between the generated answers.

\begin{table*}[t]
    \centering
    \begin{tabularx}{\textwidth}{|p{1.8cm}|X|p{4.7cm}|p{5.3cm}|p{1cm}|}
        \hline
        \rule{0pt}{2ex}\textbf{Question (Cat. / Skill)} & \textbf{Expected Response} & \textbf{RAG Benchmark} & \textbf{Ivy} & \textbf{SBERT for Ivy} \\
        \hline
        \rule{0pt}{2ex}What is the first step in solving a block world problem using means-end analysis? (Method / Means-End Analysis) & The first step in solving a block world problem using means-end analysis is to compare current arrangement to goal arrangement and determine the delta or differences in arrangements.&The first step in solving a block world problem using means-end analysis is to compare the current state with the goal state and identify the differences between them. & The first step in solving a Block World problem using Means-End Analysis is to compare the initial block arrangement with the goal arrangement. This involves identifying the differences, or “delta," to determine what changes are needed to achieve the goal state. This comparison lays the groundwork for generating valid block moves to transform the initial arrangement into the desired one. & 0.92 \\
        \hline
        \rule{0pt}{2ex}What is the goal of the painting task in partial order planning? (Task / Partial Order Planning) & The goal of the painting task in partial order planning is to achieve the state where both the ceiling and the ladder are painted. This requires the agent to plan actions in an order that ensures both goals are met without conflicts—i.e., to first paint the ceiling and then paint the ladder. & The goal of the painting task in partial order planning is to achieve two separate objectives: painting the ceiling and painting the ladder. The challenge lies in the fact that these goals are in conflict because painting the ladder first makes it wet, preventing the robot from climbing it to paint the ceiling. Therefore, the goal is to determine the correct sequence of actions to achieve both objectives efficiently, which involves painting the ceiling first, descending the ladder, and then painting the ladder. & The goal of the painting task in partial order planning is to achieve the end state where both the ladder and the ceiling are painted, represented as “Painted(Ladder) \&  Painted(Ceiling)." In this scenario, a conflict arises: if the robot paints the ladder first, it becomes wet and cannot be used to reach the ceiling. Partial order planning helps resolve this by organizing actions to avoid such conflicts. The optimal sequence is to first paint the ceiling, then climb down, and finally paint the ladder, ensuring that the actions for each goal do not interfere with one another. & 0.89 \\
        \hline
    \end{tabularx}
    \caption{Comparison of final responses from RAG Benchmark and Ivy systems to the expected responses.}
    \label{tab:benchmark}
\end{table*}

\subsection{Semantic Similarity Results}
Ivy’s responses achieved an average SBERT score of \textbf{0.82}, with scores ranging from 0.66 to 0.93.
These results demonstrate a high level of semantic alignment.
Lower scores were primarily attributed to Ivy’s inclusion of explanatory examples, which, while enhancing learner understanding, differed from the concise phrasing of the expected answers, thereby lowering the measured similarity. Pedagogically, Ivy’s responses often surpassed the expected answers by offering relatable, learning-focused explanations. This added significant educational value, particularly for supporting skill-based learning in online environments.

\subsection{Automated Analysis by LLM judges}
Our evaluation showed that Ivy correctly identified 30 questions as irrelevant and successfully used the correct TMK files for \textbf{90\%} of the relevant questions.
On average, \textbf{83\%} of the content in the intermediate responses came directly from the TMK files, with a standard deviation of 8.8\%.
Most additions involved logical elaborations on TMK content rather than factual errors.
The second judge found that \textbf{72\%} of the content from the intermediate responses was preserved in the final responses, balancing clarity with content retention.

The results validate that Ivy strongly aligns with its design principles, effectively using TMK files to generate and refine responses.
There were two misclassifications cases which suggests the need for finer categorization of questions, especially those that span multiple TMK components.
Future work will aim to refine these classifications and expand evaluations to handle more diverse user queries, enhancing Ivy’s reliability in practical settings.

\section{Discussion}
\subsection{Comparative Analysis of RAG Benchmark and Ivy Responses}
To demonstrate how Ivy generates responses that are teleologically and causally informed compared to the RAG benchmark, we compare the final responses generated by both approaches for a ‘Method’ question and a ‘Task’ question, shown in Table \ref{tab:benchmark}.
The RAG Benchmark responses demonstrate a baseline level of accuracy by correctly identifying the key steps and goals. 
However, they often lack the depth and precision expected in the context of course-specific skill terminology and associated examples.
For example, while the RAG response correctly outlines the goal for the `painting' task in partial order planning, it does not include the logical representation of goal state that was shown in lecture content.

Conversely, Ivy's explanations are notably comprehensive and contextually relevant, going beyond correctness by incorporating precise language and structured details that align closely with course content.
In the Means-End Analysis Method question, Ivy's response not only explains the process of identifying the ``delta'' but also emphasizes its role in determining valid block moves.
Similarly, for the painting task, Ivy's explanation uses logical representations like ``Painted(Ladder) \& Painted(Ceiling)'' to explicitly communicate the goal state. 

Higher SBERT scores (e.g., 0.92 and 0.89) for Ivy's responses reflect their alignment with the developer-expected answers, validating the pedagogical superiority of the TMK-enhanced Ivy's approach over the more generic RAG Benchmark responses. 
In Appendix \ref{appendix-knowledge-trace-analysis}, we delve deeper into Ivy's knowledge trace (intermediate steps) that resulted into improved final responses for the two questions in Table \ref{tab:benchmark}.

\subsection{Validation of Research Hypotheses}
Our findings strongly support both research hypotheses (RH1 and RH2), demonstrating that Ivy, grounded in the TMK framework, effectively facilitates skill-based learning by introspectively analyzing and clearly communicating skill designs.

Ivy’s ability to explain how a skill functions (RH1) is validated through multiple evaluations, including developer voting, semantic similarity analysis, and final response comparisons. 
These evaluations consistently show Ivy’s superiority over the RAG benchmark.
Developer preferences, combined with an average SBERT score of 0.82, highlight Ivy’s ability to deliver detailed and accurate explanations that align well with expectations.
This is further demonstrated by Ivy’s detailed responses, such as its ability to identify and explain the role of ``deltas'' in the Means-End Analysis skill—an aspect notably missing from the RAG benchmark responses.

Therefore, these findings validate RH1 by demonstrating Ivy’s ability to provide comprehensive and pedagogically valuable explanations.
It goes beyond surface-level definitions, delivering deeper insights into skills and enhancing the learning experience.
The automated knowledge trace evaluation using LLM judges further supports RH2, confirming that Ivy proficiently inspects skill designs and effectively utilizes TMK files to construct responses.
Ivy accessed the relevant files in 90\% of cases, with the majority of content in the initial responses directly drawn from these files.
This structured use of TMK files strengthens Ivy’s ability to effectively link tasks to methods and capture causal relationships, validating RH2’s assertion that the TMK framework enables detailed and accurate skill inspection.

\section{Ethics Statement}
This research follows ethical AI principles, emphasizing transparency, user-centered design, and responsible use of AI.
Six TMK models were manually developed from six lessons as part of an online AI course taught at Georgia Institute of Technology, excluding sensitive information.
The use of Generative AI, particularly GPT-4-based models, was restricted to educational settings with safeguards against out-of-scope responses.

Evaluations were conducted internally using human-centric and automated methods, with no external participants or sensitive data.
Future evaluations involving learners will follow university IRB guidelines, ensuring informed consent and data protection.

\section{Conclusion}
We proposed a hybrid approach to answering skill-based learning questions by integrating structured KBAI representations using TMK models with Generative AI.
This enabled Ivy to generate logical, structured, and contextually relevant explanations that enhance skill-based learning.
Across multiple evaluations, Ivy significantly outperformed a RAG-based agent, aligning with educational goals to deepen learner engagement and comprehension.
Future work will focus on deploying Ivy in real-world settings to assess its impact and scalability.

\subsubsection{Limitations}
The manual creation of TMK models required approximately seven hours per model.
This included understanding the lesson (1–2 hours), drafting the initial model (1–2 hours), revising based on peer feedback (1 hour), and finalizing (1 hour).
We aim to automate TMK model creation and establish metrics to improve skill representation accuracy and reduce development time.
Currently, Ivy addresses general procedural questions but struggles with episodic queries about specific problem instances.
Adapting TMK models to handle these queries is a focus for future research.
Furthermore, as evaluations were conducted internally, future focus group studies with actual learners will validate Ivy’s utility and identify areas for improvement.

\section{Acknowledgments}
We are grateful to Spencer Rugaber at Georgia Tech's Design Intelligence Laboratory for his invaluable insights into TMK models and modeling.
This research has been supported by NSF Grants \#2112532 and \#2247790 awarded to the National AI Institute for Adult Learning and Online Education.

\bibliography{aaai25}

\appendix
\section{Appendix}

\subsection{Estimate Question Complexity via Prompting}
\label{appendix:question-complex-prompt}
Ivy assesses the complexity and level of detail expected by the learner's question using the following prompt:
\begin{quote}
\textbf{Prompt to Ivy:} \textit{``I will give you a question and you must assign an integer value between 1 and 4 for that question. To assign an integer value to the question follow the rules given here. 
1 \textrightarrow The question contains 3 to 5 words. 
It is very direct with one word in the question that the answer needs to address. 
The question has statements like, `answer in a sentence', `answer as briefly as possible', `give the shortest answer possible' and so on. Example: What is ED?. 2 \textrightarrow The question might contain a request for a `short' answer, `brief' answer or `few sentences' and so on. 
The question does not ask for any detail and seems straight forward. Example: Explain very briefly how you find matches?. 3 \textrightarrow The question will contain a request for `short paragraph answer' or `answer in a paragraph'. It could also contain words like `explain', `explain briefly', `elaborate' and so on. Example: Explain how you find matches? or How do you generate response? Give answer in a paragraph. 4 \textrightarrow The question will contain words like `in detail', `completely', `as much detail as possible' and so on. T
he question will have a statement or part of the question will be making a request for a very elaborate and detailed answer. Example: Explain in great detail your match making process. Return only the integer value assigned to the question.''}
\end{quote}
\begin{quote}
\textbf{User:} \textless \textit{actual question asked by the learner}\textgreater
\end{quote}

\subsection{Verification Questions}
\label{appendix:verification}
Below is the list of verification questions used in our evaluation based on five question categories.

\subsubsection{Task:}
\begin{itemize}
    \item What is the goal of the painting task in partial order planning?
    \item What are the inputs needed to prove a logical statement using resolution theorem proving?
    \item What condition must be true before classifying objects?
    \item What input is required to develop a concept definition?
    \item What is the goal of solving a block world problem with means-end analysis?
    \item What condition must be satisfied on both banks for the task to be considered safe?
\end{itemize}

\subsubsection{Knowledge:}
\begin{itemize}
    \item What is variabilization in the context of incremental concept learning?
    \item What is a subclass in a concept hierarchy?
    \item What does “no goal clobbering'' mean in the context of partial order planning?
    \item What is the purpose of calculating the delta in means-end analysis?
    \item What is a literal in the context of logical sentences?
    \item What is a “configuration" in the context of the Guards and Prisoners problem?
\end{itemize}

\subsubsection{Method:}
\begin{itemize}
    \item What is required to map percepts to equivalence classes?
    \item What happens after the boat crosses the river in the Guards and Prisoners problem?
    \item What is the purpose of checking for a contradiction in resolution theorem proving?
    \item What are the key steps in incremental concept learning?
    \item What is the first step in solving a block world problem using means-end analysis?
    \item How does the method handle conflicts between subgoals when creating the plan?
\end{itemize}

\subsubsection{Student:}
\begin{itemize}
    \item How should I modify the concept diagram if a negative example of “foo” is introduced? What about positive?
    \item What are the common features shared by eagles, bluebirds, and penguins?
    \item Can you help me remember the terms modus ponens and modus tollens?
    \item How do I represent a goal state that involves multiple actions or conditions in propositional logic?
    \item What makes a state ``productive'' in addition to being legal?
    \item How should I interpret Move (C, Table) and what does it mean in terms of block position?
\end{itemize}

\subsubsection{Cannot Answer:}
\begin{itemize}
    \item How do you make a quesadilla?
    \item In what galaxy is Earth located?
    \item Who is your favorite superhero?
    \item Why do colorless green ideas sleep furiously?
    \item Who is the president of the United States?
    \item Shall I compare thee to a summer’s day?
\end{itemize}

\subsection{Results by Question Category}
\label{appendix:results-category}
The evaluation covered five question categories: Task, Method, Knowledge, Student, and Cannot Answer. The performance of each backend by question category is shown below in Table \ref{tab:results_question_category}.

\begin{table}[H]
\centering
\begin{tabular}{|p{0.35\linewidth}|p{0.25\linewidth}|p{0.2\linewidth}|}
\hline
\rule{0pt}{2ex} \textbf{Question Category} & \textbf{RAG Benchmark} & \textbf{Ivy} \\ \hline
\rule{0pt}{2ex}Task & 12 & 21  \\ \hline
\rule{0pt}{2ex}Method & 15 & 22  \\ \hline
\rule{0pt}{2ex}Knowledge & 19 & 19  \\ \hline
\rule{0pt}{2ex}Student & 16 & 20  \\ \hline
\rule{0pt}{2ex}Cannot Answer & 13 & 33  \\ \hline
\end{tabular}
\caption{Evaluator votes by question category.}
\label{tab:results_question_category}
\end{table}

In the \emph{Task} category, Ivy received 21 votes, outperforming the RAG Benchmark which received 12 votes.
This indicates Ivy's superior ability to provide clear and accurate explanations regarding the goals of tasks.
Similarly, Ivy led in the \emph{Method} category with 22 votes, reflecting its proficiency in explaining the steps or procedures involved in completing a task.
The \emph{Knowledge} category was closely contested, with both Ivy and RAG Benchmark receiving 19 votes.
This result is logical as these questions simply target background information and definitions.
In the \emph{Student} category, Ivy maintained its lead with 20 votes, indicating its better contextual understanding of user-specific queries.
Finally, Ivy dominated the \emph{Cannot Answer} category with 33 votes, highlighting its ability to more gracefully handle out-of-scope or irrelevant questions.

\begin{table}[H]
\centering
\begin{tabular}{|p{0.51\linewidth}|p{0.12\linewidth}|p{0.12\linewidth}|}
\hline
\rule{0pt}{2ex}\textbf{Skill} & \textbf{RAG Benchmark} & \textbf{Ivy}  \\ \hline
\rule{0pt}{2ex}Classification & 13 & 19  \\ \hline
\rule{0pt}{2ex}Incremental Concept Learning & 11 & 21  \\ \hline
\rule{0pt}{2ex}Means-End Analysis & 10 & 20  \\ \hline
\rule{0pt}{2ex}Planning & 14 & 14  \\ \hline
\rule{0pt}{2ex}Resolution Theorem Proving & 15 & 23  \\ \hline
\rule{0pt}{2ex}Semantic Networks & 12 & 18  \\ \hline
\end{tabular}
\caption{Evaluator votes by modeled skill.}
\label{tab:results_skill}
\end{table}

\begin{table*}[h!]
\centering
\resizebox{\textwidth}{!}{%
\begin{tabular}{|m{2.5cm}|m{5cm}|m{3.5cm}|m{5cm}|}
\hline
\rule{0pt}{2ex}\textbf{Metric} & \textbf{Description} & \textbf{Desired Outcome} & \textbf{Notes} \\ \hline
\rule{0pt}{2ex}Correctness & The accuracy and validity of the response generated by the AI agent. & High correctness & A response with high correctness should be \textbf{factually accurate} (based on TMK) to the question or context. \\ \hline
\rule{0pt}{2ex}Completeness & The response fully addresses the user’s query. & High completeness & A response with high completeness \textbf{satisfactorily covers all aspects} of a user’s query, ensuring no critical information is left out. \\ \hline
\rule{0pt}{2ex}Confidence & The degree of certainty the AI agent has regarding the accuracy or appropriateness of its answer to the user’s query. & High confidence & A response with high confidence is indicated by \textbf{straightforward, factual answers}, while terms like 'not sure,' 'likely,' or 'could be' signify medium to low confidence. \\ \hline
\rule{0pt}{2ex}Comprehensibility & The ease with which a user can understand the response generated by the AI agent. & High comprehensibility & A response with high comprehensibility is \textbf{easy to understand, useful}, and \rule{0pt}{2ex}\textbf{actionable}. Reduces the likelihood of misunderstandings or need for follow-up questions. \\ \hline
\rule{0pt}{2ex}Compactness & The quality of conveying necessary information in a concise and efficient manner. & High compactness & A response with high compactness is \textbf{clear and to the point}, without unnecessary elaboration. \\ \hline
\end{tabular}%
}
\caption{Evaluation metrics for AI generated responses.}
\label{tab:xai-metrics}
\end{table*}

\subsection{Results by Skill}
\label{appendix:results-skill}
The evaluation also examined performance by skill/problem across six different areas: Classification, Incremental Concept Learning, Means-End Analysis, Planning, Resolution Theorem Proving, and Semantic Networks. The results are presented below in Table \ref{tab:results_skill}.

For \emph{Classification}, Ivy received 19 votes, outperforming both RAG Benchmark which received 13 votes.
In \emph{Incremental Concept Learning}, Ivy was the clear leader with 21 votes, compared to 11 for RAG Benchmark.
In \emph{Means-End Analysis}, Ivy scored 20 votes, while RAG Benchmark had 10 votes. 
\emph{Planning} was the only category where Ivy and RAG Benchmark tied with 14 votes each. 
Ivy also excelled in \emph{Resolution Theorem Proving}, receiving 23 votes, surpassing RAG Benchmark's 15 votes.
Lastly, in the \emph{Semantic Networks} category, Ivy led with 18 votes, while RAG Benchmark had 12 votes.

The results show that Ivy performed almost consistently better across multiple skills, particularly in more complex areas like Incremental Concept Learning and Resolution Theorem Proving. 

\subsection{Human-centered Metrics}
\label{appendix:xai-metrics}
Table \ref{tab:xai-metrics} outlines the five human-centric metrics employed to evaluate IVY's response quality. This provides a comprehensive framework for AI-generated responses from a user-focused perspective. \\
The Completeness metric, analogous to “Recall" in information retrieval, differs from the traditional quantitative approaches by adopting a user-focused and qualitative lens to assess how thoroughly a response covers a question.  Similarly, the Confidence metric goes beyond probabilistic expression. It evaluates the response's factual nature such as directness of the answers and the absence of terms such as “not sure" or “likely". 

\subsection{Ivy Response Generation Steps and Knowledge Trace}
\label{appendix-knowledge-trace-analysis}
\subsubsection{Task Response Steps:} This is an example response from the evaluation of the Partial Order Planning skill.
Ivy generates an initial response from what it deems are the top 3 most relevant TMK parts to the user's question. 
Then it then refines this response by reorganizing and removing repeated or irrelevant text.
This cuts down the word count by roughly 50\%.  \\\\
\textbf{User Question:} What is the goal of the painting task in partial order planning?  \\\\
\textbf{Knowledge Retrieval Module:} Analysis showed a k-score of 3 meaning 3 most relevant TMK parts which were the 'paint ladder and ceiling' task, and 'Robot Tasked with Painting Problem' knowledge, and 'partial order planning' knowledge docs. \\\\
    \textbf{Response Generation Module:} Output of this module: \emph{“The goal of the painting task in partial order planning is to achieve the desired end state where both the ladder and the ceiling are painted. Specifically, the goal state is represented as “Painted(Ladder) \& Painted(Ceiling)," meaning that both the ladder and the ceiling must be in a painted condition.In this scenario, the two goals are in conflict: if the robot paints the ladder first, it will become wet, preventing the robot from climbing on it to paint the ceiling. Therefore, the planning process must account for this conflict by creating a sequence of actions that allows for both goals to be achieved without interference. Partial order planning, also known as nonlinear planning, is particularly useful in this context as it helps avoid conflicts in advance to achieve multiple goals. To reach the goal, the robot must first paint the ceiling, then climb down, and finally paint the ladder. This involves organizing the plans in such a way that the actions taken to achieve one subgoal do not interfere with the actions needed for the other subgoal. By following this structured approach, the robot can successfully complete the overall painting task without clobbering the necessary actions for each goal."}\\\\
    \textbf{Response Optimizer Module:} The generated response is refined to enhance clarity and conciseness. In this example, it retains the essential explanation of the goal state and its representation as “Painted(Ladder) \& Painted(Ceiling)" for clarity. Redundant details, such as excessive elaboration on partial order planning as nonlinear planning, are removed to streamline the response. Additionally, the explanation of conflict resolution is simplified to focus on the logical sequence of actions, ensuring the final response remains accurate and concise. This refinement reduces the response length by approximately 50\% while maintaining its relevance to the user's question. \\\\
    \textbf{Final Response:} \emph{“The goal of the painting task in partial order planning is to achieve the end state where both the ladder and the ceiling are painted, represented as “Painted(Ladder) \&  Painted(Ceiling)." In this scenario, a conflict arises: if the robot paints the ladder first, it becomes wet and cannot be used to reach the ceiling. Partial order planning helps resolve this by organizing actions to avoid such conflicts. The optimal sequence is to first paint the ceiling, then climb down, and finally paint the ladder, ensuring that the actions for each goal do not interfere with one another."}

\end{document}